\documentclass[sigconf]{acmart}

\AtBeginDocument{%
  }

\usepackage{amsmath,amsfonts}
\usepackage{tabularx}
\usepackage{subcaption}
\usepackage{multirow}
\usepackage{array} 
\usepackage{booktabs}
\usepackage[capitalise]{cleveref}
\usepackage{makecell}
\usepackage{balance}

\copyrightyear{2026}
\acmYear{2026}
\setcopyright{cc}
\setcctype{by}
\acmConference[ISWC '26]{Proceedings of the 2026 ACM International Symposium on Wearable Computers}{October 11--15, 2026}{Shanghai, China}
\acmBooktitle{Proceedings of the 2026 ACM International Symposium on Wearable Computers (ISWC '26), October 11--15, 2026, Shanghai, China}
\acmDOI{10.1145/3830727.3834824}
\acmISBN{979-8-4007-2872-3/2026/10}

\begin{document}

\title{Rotation-Invariant Multi-IMU Activity Recognition under Independent Per-Location Orientation Shifts}

\author{Seungyeol Baek}
\affiliation{%
  \institution{Korea University}
  \city{Seoul}
  \country{Republic of Korea}}
\email{mbaek01@korea.ac.kr}

\author{Yoonbyung Chai}
\affiliation{%
  \institution{Korea University}
  \city{Seoul}
  \country{Republic of Korea}}
\email{yoonbyung-chai@korea.ac.kr}

\author{Yonghyeon Lee}
\affiliation{%
  \institution{Massachusetts Institute of Technology}
  \city{Cambridge}
  \state{Massachusetts}
  \country{USA}}
\email{yhlee.gabe@gmail.com}

\author{Sungjoon Choi}
\affiliation{%
  \department{Dept. of Artificial Intelligence}
  \institution{Korea University}
  \city{Seoul}
  \country{Republic of Korea}}
\email{sungjoon-choi@korea.ac.kr}

\author{Sungho Suh}
\affiliation{%
  \department{Department of Artificial Intelligence}
  \institution{Korea University}
  \city{Seoul}
  \country{Republic of Korea}}
\email{sungho\_suh@korea.ac.kr}



\renewcommand{\shortauthors}{
Seungyeol Baek, Yoonbyung Chai, Yonghyeon Lee,
Sungjoon Choi, \& Sungho Suh
}
\begin{abstract}
Human Activity Recognition (HAR) with self-administered wearables, such as at-home rehabilitation and exercise monitoring, often requires reattaching inertial measurement units (IMUs) across sessions. In multi-IMU settings, this can induce independent orientation offsets across body locations, a deployment shift that conventional scalar HAR models do not structurally handle. Existing remedies rely on rotation augmentation, whose robustness depends on sampled transformations, or calibration and orientation-normalization pipelines requiring additional reference-frame assumptions or explicit procedures. We present Truly Rotation-Invariant HAR (TRI-HAR), a rotation-invariant framework that makes robustness to independent per-location IMU orientation offsets a structural model property. TRI-HAR reshapes accelerometer and gyroscope streams into triaxial vectors, applies a shared SO(3)-equivariant backbone and invariant projection to each IMU location, and fuses the resulting invariant features for activity classification. Across four multi-IMU benchmarks, TRI-HAR preserves macro-F1 under fixed independent per-location SO(3) rotations and outperforms rotation-augmented baselines under this target shift without requiring rotational augmentation. 

\end{abstract}




\maketitle

\section{Introduction}

\begin{figure}[!t]
    \centering
    \includegraphics[width=\columnwidth]{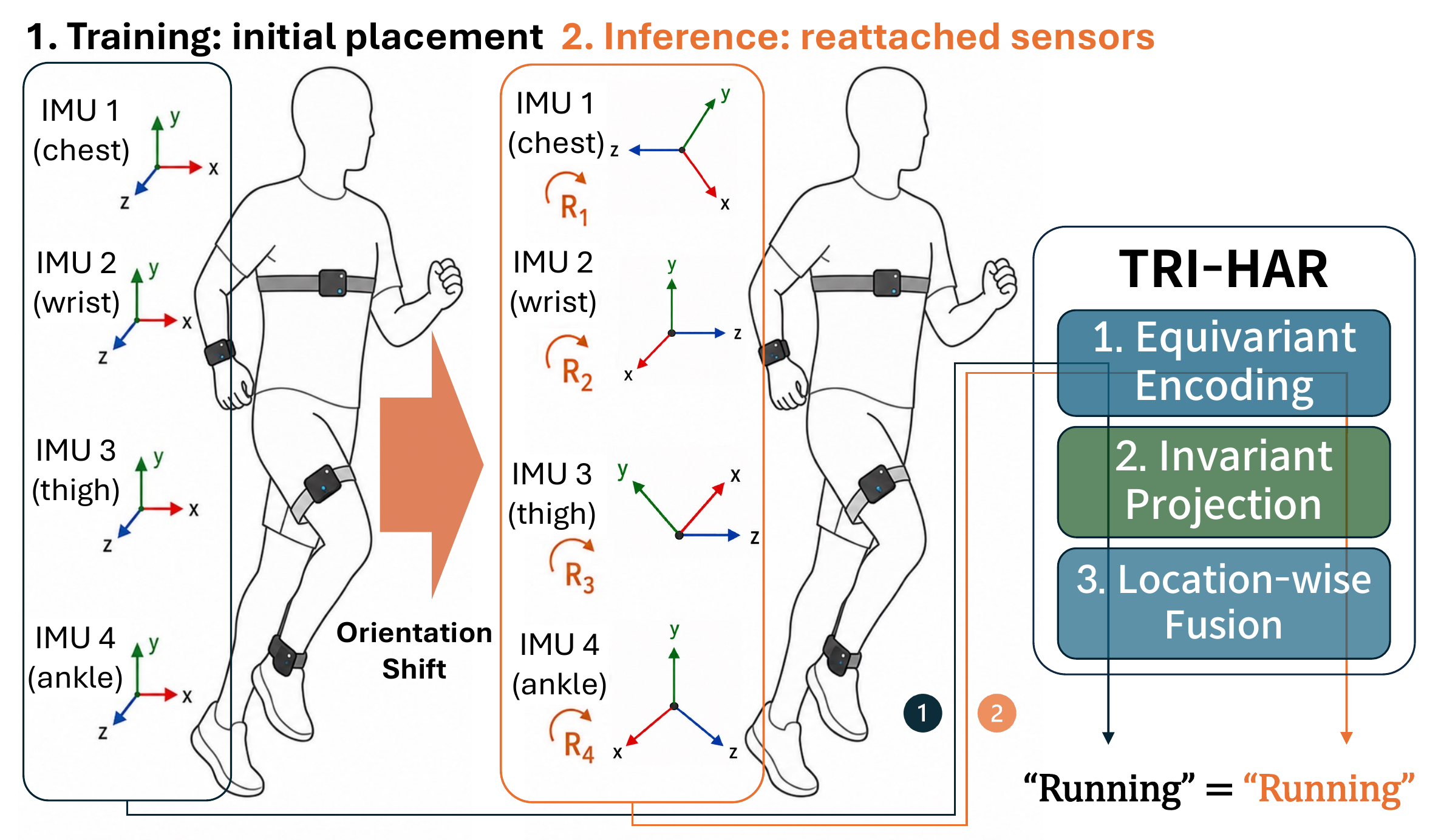} 
    \caption{Body-worn IMUs may be reattached with independent orientation offsets across uses. TRI-HAR performs per-location equivariant encoding and invariant projection before fusion for stable activity predictions.}
    \Description{Teaser diagram for TRI-HAR. The left side shows a person wearing IMUs at the chest, wrist, thigh, and ankle during an initial placement. The middle shows the same activity during inference after sensors have been reattached, with each IMU coordinate frame rotated differently and labeled with independent rotations R1 through R4. The right side shows a simplified TRI-HAR block with equivariant encoding, invariant projection, and location-wise fusion, producing the same activity label before and after reattachment.}
    \label{fig:teaser}
    \vspace{-5mm}
\end{figure}

Human Activity Recognition (HAR) classifies human activities from wearable inertial sensor data and supports applications such as healthcare monitoring, sports and exercise analytics, and gesture-based interaction \cite{de2015multimodal,muller2024imu,chen2021deep}. 
A central challenge in wearable HAR is the distribution shift caused by variations in sensor orientation, or rotational misalignment. When an IMU is worn or reattached in varying orientations, the same body motion can produce different accelerometer and gyroscope measurements. These orientation shifts change the representation of sensor signals and can substantially degrade deep HAR models that lack built-in robustness to rotation-induced distribution shift \cite{barcelo2019self,haresamudram2025past}. This problem becomes especially relevant in self-administered multi-IMU settings, such as at-home rehabilitation or exercise monitoring, where sensors may be reattached across sessions, and each body-worn IMU can acquire its own orientation offset \cite{muller2024imu}. 

Existing approaches to orientation variability typically rely on either data augmentation or calibration and orientation-normalization pipelines \cite{halmich2025data, gil2023reducing}. Rotation augmentation exposes a model to synthetic variants of the training data, but its robustness is tied to the sampled transformations and does not guarantee invariance to arbitrary unseen rotations \cite{huang2023sensor, caramaschi2023device, han2021gravity, yurtman2017activity}.
Calibration-based pipelines can reduce orientation mismatch, but they add preprocessing overhead and introduce additional assumptions about available reference frames, static intervals, magnetic reliability, or explicit user or system procedures \cite{yu2022data, halmich2025data, gil2023reducing}.
These assumptions may be difficult to satisfy in uncontrolled self-administered test settings \cite{barcelo2019self, batista2010accelerometer}.

A more structural approach is to encode rotational symmetry directly into the model. SO(3)-equivariant architectures produce intermediate features that co-rotate predictably with the input, enabling rotation-invariant representations to be constructed by design \cite{deng2021vector,son2024intuitive}. However, wearable HAR requires more than shared-global rotation invariance. In self-administered multi-IMU settings, sensors at different locations can be reattached with their own orientation offsets. 

We present Truly Rotation-Invariant HAR (TRI-HAR), a rotation-invariant framework for multi-IMU HAR under independent per-location orientation shifts. As shown in \Cref{fig:teaser}, TRI-HAR treats accelerometer and gyroscope measurements as triaxial vector streams, groups them by physical IMU location, and maps each group through a shared equivariant-to-invariant location encoder. This encoder lifts each group into a multi-frequency SO(3)-equivariant representation and projects it to an invariant location feature. The resulting invariant location features are then concatenated in a fixed body-location order for classification. This design preserves separate rotation actions until invariant projection, enabling invariance to independent per-location SO(3) offsets without rotational augmentation or an additional test-time orientation-calibration step.

We evaluate TRI-HAR on four public multi-IMU HAR benchmarks: PAMAP2, DSADS,
Opportunity, and RealDISP. Against supervised HAR baselines with and without matched per-location rotation augmentation, TRI-HAR preserves macro-F1 under fixed independent SO(3) test rotations and outperforms rotation-augmented baselines under this target shift. We also measure live host-side latency on 3-IMU and 5-IMU streams to characterize the runtime cost of the location-wise encoder.

Our contributions are summarized as follows:
\begin{itemize}
    \item We propose TRI-HAR, a multi-IMU HAR framework that learns equivariant-to-invariant representations per physical IMU location, enabling structural robustness to independent orientation offsets without rotation augmentation or additional test-time orientation calibration.
    \item We formulate an independent per-location SO(3) shift protocol, where each IMU location can receive a distinct fixed orientation offset, matching self-administered multi-IMU reattachment settings.
    \item Across four public benchmarks, we show that TRI-HAR preserves macro-F1 under fixed independent per-location rotations and outperforms rotation-augmented baselines under this target orientation-shift condition.
\end{itemize}

\section{Related Work}
Prior HAR work mainly handles orientation variability through data augmentation, ranging from simple signal-space rotations to synthetic IMU generation methods \cite{yu2022data,halmich2025data}. More recent cross-dataset approaches extend this idea with large-scale representation learning; for example, CrossHAR \cite{hong2024crosshar} combines hierarchical self-supervised pretraining with physics-informed augmentation, while oneHAR \cite{wei2025one} uses LLM-assisted virtual IMU generation to improve generalization across sensor positions and orientations.

However, these methods remain sampling-dependent. Most HAR augmentation schemes apply only restricted families of rotations, often around gravity or body axes, so robustness depends on the synthesized transformations seen during training rather than being guaranteed for arbitrary SO(3) rotations \cite{han2021gravity,caramaschi2023device,yurtman2017activity}. Calibration, canonicalization, and orientation-normalization pipelines provide an alternative by mapping measurements to a consistent reference frame \cite{tedaldi2014robust,yurtman2018activity,gil2023reducing}. While often effective within distribution, such preprocessing introduces additional engineering assumptions, such as static intervals or reliable magnetic references, that may not hold in practical wearable deployments \cite{tedaldi2014robust,yurtman2018activity,batista2010accelerometer}.

A complementary line of work encodes SO(3)-equivariance directly in the network architecture. Earlier examples include Tensor Field Networks and 3D Steerable CNNs \cite{thomas2018tensor,weiler20183d}. More directly relevant to our setting, Vector Neurons (VN) lift scalar features to 3D vectors and redefine standard neural-network operations so that they remain SO(3)-equivariant, enabling equivariant variants of backbones such as PointNet and DGCNN \cite{deng2021vector,qi2017pointnet,wang2019dynamic}. FER extends VN with a higher-dimensional multi-frequency equivariant embedding that captures richer angular structure while preserving strict rotation equivariance \cite{son2024intuitive}. TRI-HAR builds on this FER-VN line and adapts it from 3D geometric data to windowed IMU sequences for rotation-invariant HAR.

\begin{figure*}[t!]
    \centering
    \includegraphics[width=\textwidth]{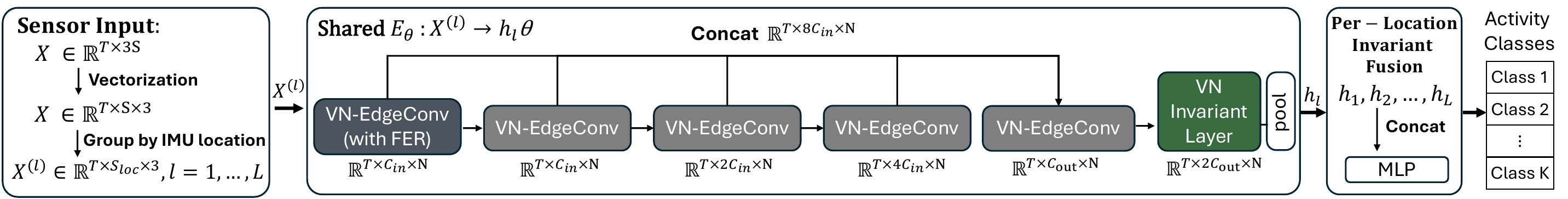} 
    \Description{Diagram of the TRI-HAR architecture. A sensor input window is first vectorized from flattened accelerometer and gyroscope channels into triaxial streams, then grouped by IMU location. Each location group is passed through the same shared encoder, labeled $E_\theta$, which contains VN-EdgeConv layers, a VN invariant layer, and pooling, producing one invariant feature $h_l$ per location. The invariant features from all locations are concatenated and passed through an MLP to produce activity-class predictions.}
    \caption{Architecture of TRI-HAR. Windowed input signals are vectorized into triaxial streams and grouped by physical IMU location. A shared location encoder $E_\theta$ is applied to each group $X^{(l)}$ to produce an invariant feature $h_l$; the features $h_1,\ldots,h_L$ are then concatenated in a fixed location order and passed to an MLP classifier. Dimensions inside $E_\theta$ are shown for one location group, and the encoder weights are shared across locations.}
    \label{fig:model}
    \vspace{-5mm}
\end{figure*}

\section{Method}
\label{sec:method}
Given a window of accelerometer and gyroscope signals, TRI-HAR reshapes scalar channels into triaxial vector streams, groups them by physical IMU location, and maps each group through a shared $\mathrm{SO}(3)$-equivariant backbone followed by invariant projection. The backbone lifts the streams into a multi-frequency angular Vector Neuron representation, and the invariant projection converts the resulting equivariant features to a rotation-invariant location feature~\cite{deng2021vector,son2024intuitive}. These invariant location
features are then concatenated in a fixed order for activity classification. 
\Cref{fig:model} summarizes the TRI-HAR architecture.
By applying invariant projection before cross-location fusion,
TRI-HAR matches self-administered wearable settings in which each body-worn IMU
may be reattached with its own orientation offset.

\subsection{Location-Grouped IMU Vector Representation}
A windowed IMU sample is initially represented as flattened scalar channels
$X_{\mathrm{raw}} \in \mathbb{R}^{T \times 3S}$, where $T$ is the window length
and $S$ is the number of triaxial streams. TRI-HAR reshapes these flat channels into triaxial vector streams, $X \in \mathbb{R}^{T \times S \times 3}$. In this representation, rotations are applied to the final 3D vector dimension of each stream.

The triaxial streams are then grouped by physical IMU location:
\begin{equation}
    X =
    \left(
    X^{(1)}, X^{(2)}, \ldots, X^{(L)}
    \right),
    \qquad
    X^{(l)} \in \mathbb{R}^{T \times S_{\mathrm{loc}} \times 3},
    \quad l=1,\ldots,L,
    \label{eq:location_grouped_input}
\end{equation}
where $L$ is the number of physical IMU locations and $S_{\mathrm{loc}}$ is the number of triaxial streams assigned to each location. These location groups define the units over which independent orientation offsets are modeled. Accelerometer and gyroscope streams from the same body-worn IMU are grouped together and share the same rotation action. The ordered tuple in \cref{eq:location_grouped_input} defines the fixed location order used for fusion. Thus, $S=L S_{\mathrm{loc}}$ in the grouped representation. TRI-HAR preserves body-location identity and is not designed to be invariant to arbitrary permutations of sensor locations.

We use $R_l \cdot X^{(l)}$ to denote applying $R_l \in \mathrm{SO}(3)$ to the 3D vector dimension of every triaxial stream at every time step within the window for location $l$. This notation models independent mounting offsets across physical IMU locations.

\subsection{Shared $\mathrm{SO}(3)$-Equivariant Backbone}
TRI-HAR uses a single shared $\mathrm{SO}(3)$-equivariant backbone, denoted $B$, for all physical IMU locations. We instantiate $B$ with FER-VN-DGCNN~\cite{son2024intuitive}, which combines Vector Neuron operations~\cite{deng2021vector} with Frequency-based Equivariant Feature Representation (FER). Vector Neurons replace scalar hidden features with vector-valued channels that transform predictably under 3D rotations. In the multi-frequency FER extension, a time-indexed latent feature is represented as $V_t \in \mathbb{R}^{C \times N}$, where $C$ is the number of vector-valued channels and $N$ is the lifted representation dimension.

Given the location-grouped input from \cref{eq:location_grouped_input}, the backbone maps each location group to a sequence of equivariant latent features:
\begin{equation}
    V_{1:T}^{(l)}
    =
    B
    \left(
    X^{(l)}
    \right),
    \qquad
    V_t^{(l)} \in \mathbb{R}^{C \times N},
    \quad l=1,\ldots,L.
    \label{eq:equivariant_encoder}
\end{equation}
Below, we describe the computation for one location group and omit the superscript $(l)$ when no ambiguity arises.

Let $D:\mathrm{SO}(3)\rightarrow\mathrm{SO}(N)$ denote the lifted rotation action used by FER. Under the right-action convention used here, a rotation $R \in \mathrm{SO}(3)$ acts on the latent feature as $V_tD(R)$. A learnable mapping $f$ is equivariant if 
\begin{equation}
    f
    \left(
    V_tD(R)
    \right)
    =
    f
    \left(
    V_t
    \right)
    D(R).
    \label{eq:vn_equiv}
\end{equation}
Thus, rotating the input causes intermediate features to co-rotate in the lifted feature space, rather than changing the underlying activity information.

FER performs the initial lift from 3D IMU vectors to this multi-frequency equivariant space. For an input vector $u \in \mathbb{R}^3$, FER defines
\begin{equation}
    \psi(u)
    :=
    \varphi(\|u\|)
    D\!\left(R^{\hat{z}}(\hat{u})\right)\hat{e},
    \label{eq:fer_core}
\end{equation}
where $\hat{u}=u/\|u\|$, $R^{\hat{z}}(\hat{u}) \in \mathrm{SO}(3)$ is the
rotation that aligns a reference axis $\hat{z}$ to $\hat{u}$, $\hat{e}$ is a
basis vector in the target embedding space, and $\varphi(\cdot)$ is a learnable
radial function applied to the input magnitude. The direction of $u$ determines
the angular component of the lifted feature, while $\varphi(\|u\|)$ modulates
its magnitude-dependent frequency coefficients. This gives TRI-HAR a
multi-frequency angular representation while preserving SO(3)-equivariance.

Within $B$, each location group is processed as a dynamic graph over time-indexed IMU states. The nodes are the $T$ time steps within the window, and the node feature at time $t$ is the colocated multi-stream sensor state $X_t^{(l)} \in \mathbb{R}^{S_{\mathrm{loc}} \times 3}$.
After the FER lift, each VN-EdgeConv block recomputes a $k$-nearest-neighbor graph among these time-step nodes in the current equivariant latent feature space.
Thus, the backbone relates feature-similar motion states from the same physical IMU location, rather than constructing nodes over individual scalar channels or sensor-time pairs.

Here, VN-EdgeConv denotes the Vector-Neuron implementation of the EdgeConv update used for graph feature aggregation.
For a central node feature $V_i$ and neighbor feature $V_j$, the edge message is computed as $m_{ij}=\mathrm{VN\text{-}MLP}([V_i,V_j-V_i])$, where the concatenation is along the vector-channel dimension. 
This construction combines the current node feature with its relative difference to a neighbor
while preserving equivariance through VN linear layers, pooling, and
nonlinearities. We use the norm-based VN nonlinearity from FER-VN-DGCNN, which
rescales vector features using functions of their rotation-invariant norms,
rather than the original VN-ReLU~\cite{deng2021vector,son2024intuitive}.

Stacking these VN-EdgeConv blocks yields the equivariant feature
sequence $V_{1:T}^{(l)}$ used by the invariant projection described
next.

\subsection{Rotation-Invariant Location Features}
\label{subsec:vn_invariant_projection}

The equivariant backbone produces features that co-rotate with the input. TRI-HAR converts these features into invariant scalar representations using the VN invariant layer.

Let $V_t \in \mathbb{R}^{C \times N}$ be an equivariant feature and let $F_t \in \mathbb{R}^{C' \times N}$ be a learned equivariant frame. If both transform under the same lifted rotation $D(R)$, then
\begin{equation}
    \left(V_tD(R)\right)
    \left(F_tD(R)\right)^\top
    =
    V_tD(R)D(R)^\top F_t^\top
    =
    V_tF_t^\top .
    \label{eq:vn_invar_property}
\end{equation}
This allows invariant scalars to be constructed from inner products between the latent feature and a co-rotating frame.

We first compute a window-level context $\bar{V}$. Then, for each time step
$t$, we estimate an equivariant frame $F_t$ and obtain the invariant time-step feature $z_t$:
\begin{equation}
\begin{aligned}
    \bar{V} &:= \frac{1}{T}\sum_{\tau=1}^{T} V_\tau, \\
    F_t &:= \mathrm{VN\text{-}MLP}([V_t,\bar{V}]), \\
    z_t &:= \mathrm{VN\text{-}In}(V_t) = V_tF_t^\top .
\end{aligned}
\label{eq:vn_invariant_layer}
\end{equation}
Thus, $z_t$ is invariant to the rotation of the input that produced $V_t$.

For a physical IMU location $l$, the invariant time-step features
$\{z_t^{(l)}\}_{t=1}^{T}$ are aggregated over time with global max and average pooling, denoted by $\mathrm{Pool}$, to obtain a single location-level invariant representation $h_l$. We use $E_\theta$ to denote the full shared location encoder:
\begin{equation}
    h_l :=
    E_\theta(X^{(l)}),
    \qquad
    E_\theta :=
    \mathrm{Pool} \circ \mathrm{VN\text{-}In} \circ B .
    \label{eq:shared_location_encoder}
\end{equation}
Here, $\theta$ collects the learnable parameters of the shared location encoder, including those in the equivariant backbone $B$ and the VN invariant projection. The same $E_\theta$ is used for every location group, but it is evaluated separately on each $X^{(l)}$, so TRI-HAR shares weights across locations without learning location-specific equivariant encoders.

\subsection{Per-Location Invariant Fusion}
\label{subsec:per_location_fusion}
TRI-HAR fuses information across IMU locations only after each location group has been mapped to an invariant feature. The resulting location features $h_1,\ldots,h_L$ are concatenated in a fixed dataset-specific order and passed to the final classifier:
\begin{equation}
    \hat{\mathbf{y}}
    =
    \mathrm{MLP}_\omega([h_1,h_2,\ldots,h_L]).
    \label{eq:per_location_fusion}
\end{equation}
Although location features are computed separately, the final classifier receives
their fixed-order concatenation, so activity prediction can still depend on
multi-location patterns in the invariant feature space.

This fusion order gives TRI-HAR invariance to independent per-location orientation offsets. In \cref{eq:vn_invar_property}, the invariant projection cancels the lifted rotation because both the latent feature $V_t$ and the learned frame $F_t$ transform by the same $D(R)$. Combining vector features from multiple IMU locations before invariant projection would therefore impose a shared-global rotation assumption on the joint representation. TRI-HAR instead applies invariant projection within each location group, canceling the local action $D(R_l)$ before fusing locations.

Suppose that location $l$ receives an independent rotation
$R_l \in \mathrm{SO}(3)$. Equivariance of $B$ gives $B(R_l \cdot X^{(l)}) = B(X^{(l)})D(R_l)$.
The invariant projection removes this location-specific rotation action, and pooling aggregates the invariant time-step features. Thus, the full location encoder satisfies
\begin{equation}
    E_\theta(R_l \cdot X^{(l)})
    =
    E_\theta(X^{(l)}),
    \qquad
    l=1,\ldots,L.
    \label{eq:per_location_encoder_invariance}
\end{equation}
Consequently, replacing each location input $X^{(l)}$ in \cref{eq:per_location_fusion} with its independently rotated version $R_l \cdot X^{(l)}$ leaves every $h_l$ unchanged, and therefore leaves $\hat{\mathbf{y}}$ unchanged.

This establishes invariance to independent per-location rotations, with shared
global rotation as a special case. It is also the central distinction from a
joint-fusion equivariant control, which does not preserve separate location
groups through invariant projection and therefore matches a shared-global
rotation assumption rather than the independent reattachment setting targeted
here.

\newcommand{\NA}{\multicolumn{1}{c}{\textemdash}}
\newcommand{\locfix}{\mathrm{SO}(3)_{\text{loc-fix}}}

\section{Experiments}
\label{sec:exp}

\subsection{Datasets and Preprocessing}
We evaluate TRI-HAR on PAMAP2 \cite{reiss2012introducing}, DSADS \cite{altun2010comparative},
Opportunity \cite{chavarriaga2013opportunity}, and RealDISP
\cite{banos2012benchmark}. \Cref{tab:datasets} summarizes the
sampling rate, window length, number of subjects, number of classes,
and number of IMU locations used in our experiments. We use the listed
sampling rates and window lengths following the selected baselines
\cite{zhou2022tinyhar,mahmud2020human}. For consistency, we retain accelerometer and gyroscope channels only.
For PAMAP2, we remove subject~109 due to missing data and use the
\(\pm16g\) accelerometer stream. For Opportunity, we exclude the \textsc{Null} label and retain seven on-body IMU locations; for the shoe IMUs, we use the sensor/body-frame
accelerometer and gyroscope streams and omit navigation-frame streams. For
RealDISP, we include the ideal, self-displacement, and mutual-displacement
recordings. 
Signals are segmented with 50\%-overlap sliding windows using
the listed window lengths and standardized channel-wise using
training-split statistics. 

        

\begin{table}[!t]
    \centering
    \caption{Characteristics of the datasets used in the evaluation}
    \label{tab:datasets}

    \begin{minipage}{\columnwidth}
        \setlength{\tabcolsep}{7pt}
        \begin{tabularx}{\linewidth}{@{} X c c c c c @{}}
            \toprule
            \textbf{Dataset}
            & \textbf{Freq.}
            & \textbf{WL}
            & \textbf{\#Subj.}
            & \textbf{\#Cls.}
            & \textbf{\#Locs.} \\
            \midrule
            PAMAP2 \cite{reiss2012introducing}
                & 33 & 5.12 & 8  & 12 & 3 \\
            DSADS \cite{altun2010comparative}
                & 25 & 5.00 & 8  & 19 & 5 \\
            Opportunity \cite{chavarriaga2013opportunity}
                & 30 & 1.00 & 4  & 17 & 7 \\
            RealDISP \cite{banos2012benchmark}
                & 50 & 2.40 & 17 & 33 & 9 \\
            \bottomrule
        \end{tabularx}

        \vspace{2pt}
        {\footnotesize
        \raggedright
        \textbf{Note:}
        \textit{Freq.} denotes sampling rate (Hz);
        \textit{WL} denotes window length (s);
        \textit{Subj.} and \textit{Cls.} indicate the numbers of
        subjects and activity classes, respectively; and
        \textit{Locs.} denotes the number of body-worn sensor
        locations.\par
        }
    \end{minipage}
\end{table}

\begin{table*}[t]
\centering
\caption{Benchmark comparison under fixed independent per-location test rotations. Values are macro-F1 (\%, mean \(\pm\) standard deviation) over cross-subject folds. Columns report performance on the original test data \(I\) and under \(\locfix\); entries under \(I\) are omitted for loc-sample rows for compactness. Bold denotes the best value within each dataset/test column.}
\label{tab:f1_scores}
\setlength{\tabcolsep}{3pt}
\small
\begin{tabular*}{\textwidth}{@{\extracolsep{\fill}} ll cc cc cc cc @{}}
\toprule
\multirow{2}{*}{\textbf{Model}} & \multirow{2}{*}{\textbf{Train Aug.}}
& \multicolumn{2}{c}{\textbf{PAMAP2} \cite{reiss2012introducing} }
& \multicolumn{2}{c}{\textbf{DSADS} \cite{altun2010comparative}}
& \multicolumn{2}{c}{\textbf{Opportunity} \cite{chavarriaga2013opportunity} }
& \multicolumn{2}{c}{\textbf{RealDISP} \cite{banos2012benchmark}} \\
\cmidrule(lr){3-4}\cmidrule(lr){5-6}\cmidrule(lr){7-8}\cmidrule(lr){9-10}
& & $\mathbf{I}$ & $\mathbf{\locfix}$
  & $\mathbf{I}$ & $\mathbf{\locfix}$
  & $\mathbf{I}$ & $\mathbf{\locfix}$
  & $\mathbf{I}$ & $\mathbf{\locfix}$ \\
\midrule

\multirow{2}{*}{MC-CNN \cite{yang2015deep}}
& none
& 80.45$\pm$8.18 & 31.52$\pm$6.21
& 82.80$\pm$8.09 & 30.63$\pm$5.80
& 41.18$\pm$6.62 & 11.82$\pm$3.90
& 87.48$\pm$10.39 & 48.95$\pm$9.81 \\
& loc-sample
& \NA & 79.14$\pm$6.75
& \NA & 81.75$\pm$6.43
& \NA & 33.56$\pm$8.10
& \NA & 87.19$\pm$8.35 \\
\addlinespace[2pt]

\multirow{2}{*}{DeepConvLSTM \cite{ordonez2016deep}}
& none
& 71.13$\pm$10.39 & 21.84$\pm$10.05
& 76.05$\pm$11.27 & 25.85$\pm$6.74
& 39.44$\pm$7.99 & 6.16$\pm$1.85
& 86.99$\pm$10.21 & 53.37$\pm$16.25 \\
& loc-sample
& \NA & 76.65$\pm$8.80
& \NA & 77.97$\pm$7.94 
& \NA & 30.01$\pm$9.06
& \NA & 87.65$\pm$7.65 \\
\addlinespace[2pt]

\multirow{2}{*}{MLP-HAR \cite{zhou2024mlp}}
& none
& 80.08$\pm$6.73 & 28.52$\pm$9.27 
& 83.23$\pm$7.80 & 22.55$\pm$6.46
& \textbf{42.01$\pm$3.93} & 5.45$\pm$1.32
& 88.49$\pm$6.82 & 39.70$\pm$4.80 \\
& loc-sample
& \NA & 79.67$\pm$8.74
& \NA & 86.11$\pm$6.83
& \NA & 32.67$\pm$4.61
& \NA & 88.38$\pm$5.64 \\
\addlinespace[2pt]

\multirow{2}{*}{TinyHAR \cite{zhou2022tinyhar}}
& none
& 82.10$\pm$6.52 & 31.64$\pm$4.77
& 82.82$\pm$7.25  & 26.60$\pm$10.33
& 38.13$\pm$4.54 & 7.37$\pm$1.12
& 88.13$\pm$8.78 & 52.43$\pm$8.74 \\
& loc-sample
& \NA & 79.99$\pm$7.46
& \NA & 82.50$\pm$8.01
& \NA & 29.98$\pm$5.44
& \NA &  88.36$\pm$8.71\\
\addlinespace[2pt]

\multirow{2}{*}{SA-HAR \cite{mahmud2020human}}
& none
& 74.74$\pm$10.27 & 12.65$\pm$5.55
& 78.11$\pm$7.87 & 16.14$\pm$3.32
& 28.78$\pm$9.56 & 4.84$\pm$1.15
& 81.78$\pm$16.14 & 50.54$\pm$14.29 \\
& loc-sample
& \NA & 70.59$\pm$7.20 
& \NA & 72.95$\pm$8.01 
& \NA & 21.23$\pm$6.13
& \NA & 82.51$\pm$15.26 \\ 

\midrule
TRI-HAR & none
& \textbf{83.64$\pm$5.56} & \textbf{83.64$\pm$5.56}
& \textbf{89.07$\pm$4.49} & \textbf{89.07$\pm$4.49}
& 38.06$\pm$3.77 & \textbf{38.06$\pm$3.77}
& \textbf{91.94$\pm$4.99} & \textbf{91.94$\pm$4.99} \\
\bottomrule
\end{tabular*}
\end{table*}

\subsection{Baselines}
We compare TRI-HAR with five supervised HAR baselines: MC-CNN~\cite{yang2015deep}, DeepConvLSTM~\cite{ordonez2016deep}, MLP-HAR~\cite{zhou2024mlp}, TinyHAR~\cite{zhou2022tinyhar}, and SA-HAR~\cite{mahmud2020human}. Unless noted otherwise, we use the original architectures from the cited papers, adapting only the input and output dimensions to each dataset. All baselines use the same preprocessing, cross-validation splits, and test-rotation protocol as TRI-HAR. Our comparison focuses on supervised HAR backbones under a matched independent-rotation augmentation protocol, isolating the architectural effect of replacing sampled augmentation with built-in per-location invariance. For MLP-HAR, \(\tau\) was set to evenly divide each dataset-specific window, yielding three temporal patches per window except for DSADS, which uses five one-second patches.

\subsection{Implementation Details}
\subsubsection{Batched Location-Encoder Implementation}
For efficient computation, our implementation%
\footnote{\url{https://github.com/mbaek01/TRI-HAR}}
folds the location axis into the batch axis, so location groups are evaluated by the shared encoder \(E_\theta\) in one batched forward pass. The location features are then reshaped to restore the batch axis and fixed location order before fusion. 

\subsubsection{Evaluation Protocol and Rotation Setup}
We adopted a unified cross-subject evaluation protocol across all benchmarks, with folds defined by subject or subject group. For PAMAP2 (8 folds), DSADS (8 folds), and Opportunity (4 folds), we used leave-one-subject-out cross-validation. For RealDISP, we used a 5-fold grouped split designed to evaluate robustness to sensor displacement. Four folds contain subject groups from the \textit{Ideal} and \textit{Self-displacement} scenarios, while the fifth fold contains all \textit{Mutual-displacement} samples. In this scenario, the instructor repositions and rotates the sensors to purposely introduce substantial variation in rotational orientation and asymmetric placement errors. This design ensures that the displacement and orientation variations in the \textit{Mutual-displacement} setting are not observed during training, providing a stringent test of robustness under realistic placement variability.

To evaluate robustness to orientation variability in multi-IMU settings, we consider two test conditions: the original test data, \(I\), and fixed independent per-location SO(3) offsets, \(\mathrm{SO}(3)_{\text{loc-fix}}\). In \(\mathrm{SO}(3)_{\text{loc-fix}}\), one SO(3) rotation is sampled independently for each physical IMU location and held fixed across the held-out fold. This protocol is motivated by the target deployment setting, in which different body-worn IMUs may be reattached with different mounting orientations. In contrast to prior equivariant-learning work, which commonly evaluates shared-global SO(3) augmentation \cite{esteves2018learning,deng2021vector}, our setting focuses on independent per-location offsets.

For the augmented baselines, we use a matched independent per-location SO(3) augmentation, denoted by loc-sample: for each training window, one random SO(3) rotation is sampled independently for each location group and applied jointly to all channels from that IMU location group (accelerometer and gyroscope). These per-location rotations are resampled on the fly for every training window. We use this augmentation rather than shared-global SO(3) augmentation because it better matches the target deployment condition and empirically yields stronger baseline robustness under \(\mathrm{SO}(3)_{\text{loc-fix}}\). TRI-HAR is trained without rotational augmentation and instead achieves invariance to the modeled per-location SO(3) offsets by construction.

\subsubsection{Hyperparameters}
All models were trained with cross-entropy loss and Adam \cite{kingma2014adam} (\(\beta_1=0.9, \beta_2=0.999\), learning rate \(1\times10^{-3}\)) for up to 150 epochs on an NVIDIA RTX 4090 (24GB) GPU, using batch size 128 and early stopping on validation macro-F1. For TRI-HAR, we set \(C_{\text{in}}=14\), \(C_{\text{out}}=224\). We use $k=20$ for Opportunity and $k=5$ for the remaining datasets, selected by validation macro-F1.

\subsubsection{Latency Evaluation}
Latency was measured on a Ryzen 7 9800X3D host with an RTX 5090 using native-rate ESP32-S3 streams matched to the PAMAP2-derived 3-IMU setting at 33 Hz and the DSADS-derived 5-IMU setting at 25 Hz. The ESP32-S3 streamed IMU samples only; windowing, preprocessing, and inference ran on the host. We report p99 live model and window-to-label latency over 1000 windows, where window-to-label latency is measured from final-sample arrival to label production. A stream is considered feasible for steady-state real-time operation if p99 window-to-label latency is below the window-update period \(H/f_s\). These measurements characterize live host-side streaming inference in the tested setup.

\section{Results}
\label{sec:results}

\subsection{Benchmark Comparison}
\Cref{tab:f1_scores} compares TRI-HAR with supervised HAR baselines on the original test data \(I\) and under fixed independent per-location offsets \(\locfix\). 
Without rotation augmentation, all non-equivariant baselines degrade substantially under \(\locfix\), whereas TRI-HAR preserves its macro-F1 under \(\locfix\) across all four datasets, consistent with its invariance by construction and relative invariance error below \(10^{-10}\).
Loc-sample augmentation improves baseline robustness, but TRI-HAR still achieves the highest macro-F1 under \(\locfix\) on all four datasets without rotation augmentation. Under \(I\), TRI-HAR is best on PAMAP2, DSADS, and RealDISP, while trailing the strongest scalar baseline on Opportunity.

\subsection{RealDISP Mutual-Displacement Fold}
Beyond the synthetic \(\mathrm{SO}(3)_{\text{loc-fix}}\) evaluation, we examine the held-out RealDISP \textit{Mutual-displacement} fold, where sensors are physically repositioned and reoriented. This fold is consistently among the most challenging of the five RealDISP folds. As shown in \cref{tab:realdisp}, TRI-HAR achieves the highest macro-F1,
exceeding the strongest non-augmented and rotation-augmented baselines, MLP-HAR and TinyHAR, by 7.74 and 6.80 points, respectively. Relative to the corresponding five-fold RealDISP means under \(I\), the non-augmented baselines are 10.94--31.89 macro-F1 points lower on the \textit{Mutual-displacement} fold, compared with 6.7 points for TRI-HAR.

Because \textit{Mutual-displacement} includes both orientation and positional changes, it is not an isolated test of rotation robustness.
Nevertheless, TRI-HAR's advantage suggests that making predictions invariant to the modeled per-location rotations by construction remains beneficial under this combined physical placement shift. 
Moreover, rotation augmentation is not consistently beneficial: it improves four of the five baselines by 1.4--6.9 points but decreases MLP-HAR by 2.9 points. TRI-HAR instead attains the highest macro-F1 with structural invariance and without the augmentation-induced degradation observed for MLP-HAR.

\begin{table}[t]
    \centering
    \caption{Macro-F1 (\%) on RealDISP \textit{Mutual-displacement} fold. \(\Delta_{\mathrm{aug}}\) is augmentation-induced change in baseline score. Bold marks the highest score.}
    \label{tab:realdisp}
    \small
    \setlength{\tabcolsep}{10pt}
    \begin{tabular}{@{}lccc@{}}
        \toprule
        \textbf{Model}
        & \textbf{none}
        & \textbf{loc-sample}
        & \(\boldsymbol{\Delta_{\mathrm{aug}}}\) \\
        \midrule
        MC-CNN
        & 67.95
        & 74.84
        & \(+6.89\) \\

        DeepConvLSTM
        & 67.68
        & 72.90
        & \(+5.22\) \\

        MLP-HAR
        & 77.55
        & 74.67
        & \(-2.88\) \\

        TinyHAR
        & 72.55
        & 78.49
        & \(+5.94\) \\

        SA-HAR
        & 49.89
        & 51.28
        & \(+1.39\) \\
        \midrule
        TRI-HAR
        & \multicolumn{2}{c}{\textbf{85.29}}
        & \textemdash \\
        \bottomrule
    \end{tabular}
    \vspace{-5mm}
\end{table}

\subsection{Location-wise Fusion and Host-side Latency}
To isolate the effect of location-wise invariant fusion, we compare TRI-HAR with TRI-HAR-joint, an architectural control that retains the same equivariant components but replaces
location-wise invariant fusion with one joint all-location invariant projection.
For PAMAP2/DSADS/Opportunity/RealDISP, TRI-HAR-joint loses 37.71/50.95/16.80/50.12 macro-F1 points under \(\locfix\)\ relative to \(I\), whereas TRI-HAR loses none.
A joint invariant projection can therefore cancel a shared global rotation but not independent per-location rotations. TRI-HAR avoids this failure by canceling each location-specific rotation before fusion. Under \(I\), TRI-HAR also exceeds TRI-HAR-joint by 3.76/2.25/3.61/5.20 macro-F1 points; shared-global SO(3) tests are equivalent to \(I\) for both equivariant models and are omitted.

This location-wise formulation creates TRI-HAR's main runtime tradeoff. Although the encoder is shared, the equivariant backbone and invariant projection are executed once per physical IMU location. We therefore report live latency on 3-IMU and 5-IMU streams to quantify the host-side runtime cost introduced by the location-wise encoder.

\Cref{tab:latency} reports live p99 model and window-to-label latency. Our feasibility check uses window-to-label latency. 
TRI-HAR is slower than scalar baselines, reflecting both the heavier equivariant-to-invariant encoder and its repeated per-location application. Nevertheless, its CPU window-to-label latency remains far below the window-update period.
With the implemented hops, \(H/f_s=84/33 \approx 2.55\) s for 3-IMU and \(H/f_s=62/25=2.48\) s for 5-IMU, whereas TRI-HAR's corresponding latencies are 67.61 ms and 85.78 ms. 
The update periods are therefore 37.6 and 28.9 times larger than the corresponding
p99 latencies, supporting host-side real-time streaming inference despite
the per-location computation.

\begin{table}[t]
\centering
\small
\setlength{\tabcolsep}{7pt}
\caption{Host-side live p99 latency for the 3-IMU and 5-IMU settings.
CPU and GPU entries report model/window-to-label in ms. FP32 denotes
parameter memory.}
\label{tab:latency}
\begin{tabular}{llrrr}
\toprule
Setting & Model & FP32 (MiB) & CPU (ms) & GPU (ms) \\
\midrule
3-IMU & MLP-HAR & 0.67 & 0.86/1.08 & 0.69/1.09 \\
 & MC-CNN & 23.14 & 1.79/2.02 & 0.53/0.94 \\
 & TRI-HAR & 13.40 & 67.38/67.61 & 2.58/2.97 \\
\midrule
5-IMU & MLP-HAR & 1.84 & 0.97/1.21 & 1.14/1.48 \\
 & MC-CNN & 28.46 & 2.61/2.84 & 0.88/1.25 \\
 & TRI-HAR & 21.41 & 85.56/85.78 & 3.01/3.36 \\
\bottomrule
\end{tabular}
\vspace{-5mm}
\end{table}

\subsection{Discussion and Limitations}
TRI-HAR targets calibration-light, self-administered multi-IMU HAR, such as at-home rehabilitation or exercise monitoring, where sensors may be reattached across sessions and orientation should be treated as a nuisance factor. This is reflected in the multi-sensor benchmarks, where TRI-HAR remains stable under unseen rotations while the non-equivariant baselines degrade sharply. The location-wise formulation is particularly motivated by settings in which different body-worn IMUs have independent mounting orientations. Comparisons with explicit calibration, reference-frame normalization, and handcrafted orientation-invariant pipelines are complementary and left for future work.

The latency results show that the current host-side implementation can keep up 
with live 3-IMU and 5-IMU streams, supporting off-sensor inference from streaming wearable IMUs. 
TRI-HAR nevertheless remains a robustness-first architecture rather than an edge-optimized wearable model. Its computational cost comes from two sources: the shared encoder is evaluated once per IMU location, and the encoder itself is costly, with the invariant projection as the dominant stage. 
Because the first source is the architectural cost of the targeted per-location invariance, future efficiency work should focus on reducing the encoder-internal cost while preserving the location-wise fusion. Promising directions include reducing the feature width before invariant projection, lower-rate frame estimation, graph sparsification, and distillation or simplification for equivariant GNNs \cite{ekstrom2023accelerating,li2021towards,tailor2021towards}.

\section{Conclusion}
We presented TRI-HAR, a rotation-invariant wearable HAR framework that applies a
shared SO(3)-equivariant backbone and invariant projection separately to each
physical IMU location before fixed-order fusion. Across four multi-IMU
benchmarks, TRI-HAR preserved macro-F1 under fixed independent per-location
SO(3) rotations and outperformed rotation-augmented baselines under this target shift. These results support architecture-level rotation
invariance for calibration-light, self-administered
wearable HAR, with efficient on-device deployment left for
future work.


\section*{Acknowledgments}
This work was supported by the Institute of Information \& communications Technology Planning \& Evaluation (IITP) grant funded by the Korea government (MSIT) (No. RS-2019-II190079, AI Graduate School Program (Korea University), 20\%), the IITP-ITRC (Information Technology Research Center) grant (IITP-2026-RS-2024-00436857, 30\%), and the IITP grant (No. RS-2026-25519380, 50\%).

\bibliographystyle{ACM-Reference-Format}
\balance
\bibliography{refs}

\appendix









\end{document}